\documentclass[11pt,a4paper]{article}
\usepackage[hyperref]{emnlp2020}
\usepackage{times}
\usepackage{color}
\usepackage{latexsym}

\usepackage{CJKutf8}
\usepackage{stfloats}
\usepackage{graphicx}
\graphicspath{ {./images/} }
\usepackage[table,xcdraw]{}
\usepackage{microtype}
\usepackage{booktabs}
\usepackage{xcolor}         % colors

\usepackage{soul}
\usepackage{amsmath}
\usepackage{amssymb}% http://ctan.org/pkg/amssymb
\usepackage{pifont}% http://ctan.org/pkg/pifont\textbf{}
\usepackage{amssymb}
\usepackage{multirow} %合并多行单元格的宏包
\usepackage{makecell}
\usepackage{stfloats}
\usepackage{tabularx}
\usepackage{threeparttable}
\usepackage{hyperref}

\aclfinalcopy % Uncomment this line for the final submission
\title{The Box is in the Pen: Evaluating Commonsense Reasoning in Neural Machine Translation}

\author{
  Jie He\textsuperscript{1}\thanks{\ \ Equal Contributions.}\ \ ,  
  Tao Wang\textsuperscript{2}\footnotemark[1]\ \ ,  
  Deyi Xiong\textsuperscript{1},  
  \and Qun Liu\textsuperscript{3} \\  
  \textsuperscript{1}College of Intelligence and Computing, Tianjin University, Tianjin, China \\  
  \textsuperscript{2}School of Computer Science and Technology, Soochow University, Suzhou, China \\  
  \textsuperscript{3}Huawei Noah’s Ark Lab, Hong Kong, China \\  
  \texttt{jieh@tju.edu.cn, rgwt1234@gmail.com} \\  
  \texttt{dyxiong@tju.edu.cn, qun.liu@huawei.com}
}

\begin{document}
\begin{CJK}{UTF8}{gbsn}

\maketitle
\begin{abstract}

Does neural machine translation yield translations that are congenial with common sense?
In this paper, we present a test suite to evaluate the commonsense reasoning capability of neural machine translation. The test suite consists of three test sets, covering lexical and contextless/contextual syntactic  ambiguity that requires commonsense knowledge to resolve. We manually create 1,200 triples, each of which contain a source sentence and two contrastive translations, involving 7 different common sense types. Language models pretrained on large-scale corpora, such as BERT, GPT-2, achieve a commonsense reasoning accuracy of lower than 72\% on target translations of this test suite. We conduct extensive experiments on the test suite to evaluate commonsense reasoning in neural machine translation and investigate factors that have impact on this capability.  Our experiments and analyses  demonstrate that neural machine translation performs poorly on commonsense reasoning of the three ambiguity types in terms of both reasoning accuracy ( $\leqslant60.1\%$) and reasoning consistency ($\leqslant31\%$).  We will release our test suite as a machine translation commonsense reasoning testbed to promote future work in this direction.
\end{abstract}

\section{Introduction}

Sixty years ago, the pioneering machine translation researcher and linguist Bar-Hillel published his well-known argument on the non-feasibility of general-purpose fully automatic high-quality machine translation (FAHQT) due to the inevitable requirement of world knowledge to help machine translation to infer correct translations for ambiguous words or linguistic structures \cite{bar1964demonstration}. The example that Bar-Hillel uses as an evidence for the need of commonsense knowledge in machine translation is ``The box is in the pen'', where machine translation is expected to perform reasoning on the relative sizes of ``box'' and ``pen''. Bar-Hillel also doubts that a machine, even equipped with extra-linguistic knowledge, would be able to reason with such knowledge spontaneously as human translators do \cite{bar1964demonstration,Macklovitch1995The}.

Modern natural language processing (NLP) has made tremendous progress, not only in building abundant resources to develop linguistic insights, but also in plenty of methodological practices. On the one hand, machine translation has been substantially advanced with large-scale parallel data and statistical models. Recent results even suggest that the quality of machine-generated translations is approaching professional human translators \cite{wu2016google,DBLP:journals/corr/abs-1803-05567}. On the other hand, a wide variety of efforts have been conducted to either examine the commonsense reasoning capability of neural models in natural language understanding, establish commonsense reasoning challenges or enhance neural models in commonsense reasoning \cite{Zhang2018ReCoRDBT,Talmor2018CommonsenseQAAQ,huang-etal-2019-cosmos,sap-etal-2019-social}.

Comparing with Bar-Hillel's doubts and recent progress on machine translation and commonsense reasoning, it is natural for us to ask questions: do we solve the machine translation impasse related to commonsense reasoning? Or particularly, are current neural machine translation models able to learn common sense? And if so, how much do they learn? Does neural machine translation acquire sufficient commonsense knowledge and have strong ability in commonsense reasoning to generate human-level high-quality translations?
Methodological discussion on the feasibility of FAHQT given the recent progress is far beyond the scope of this work. Instead, we focus on empirically analyzing the capability of state-of-the-art neural machine translation models in using extra-linguistic commonsense knowledge to resolve ambiguity at different linguistic levels and select correct translations after disambiguation. 
% By a thorough evaluation on this, we want to understand where we are now in machine translation commonsense reasoning and where we should probably go in the future.

% \begin{figure}[t]
% \centering
% \includegraphics[scale=0.7]{images/figure_introduction.eps}
% \caption
% {Examples of the lexical ambiguity (a), contextless syntactic ambiguity (b) and contextual syntactic ambiguity (c). English Translations in bold are correct while underlined translations are incorrect. }
% \label{figure-introduction}
% \end{figure}

\begin{figure}[t]
\setlength{\belowcaptionskip}{-0.6cm} 
\centering
\includegraphics[scale=0.62]{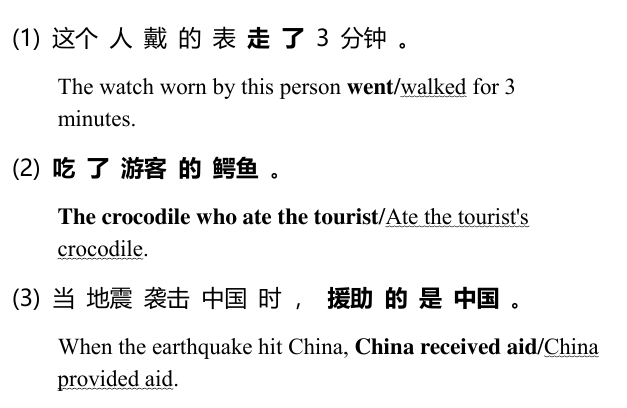}
\caption
{Examples of the lexical ambiguity (1), contextless syntactic ambiguity (2) and contextual syntactic ambiguity (3). English Translations in bold are correct while underlined translations are incorrect. }
\label{figure-introduction}
\end{figure}

In order to achieve this goal, we manually build a machine translation commonsense reasoning test suite on Chinese-to-English translation with three types of commonsense-related ambiguities: lexical ambiguity, contextless and contextual syntactic ambiguity (see Section \ref{test-suite-design} for more details). Examples are shown in Figure \ref{figure-introduction}. 
% For each ambiguous source sentence in the test suite, we provide two contrastive translations: a right and wrong translation  that only differ in the translation of ambiguous part.
With this test suite, we thoroughly evaluate the commonsense reasoning ability of state-of-the-art neural machine translation models, e.g., LSTM- and Transformer-based NMT \cite{DBLP:journals/corr/BahdanauCB14,vaswani2017attention}. We also conduct analyses on the commonsense reasoning capability according to commonsense knowledge types, sentence length and reasoning consistency and the size of training data. 

To the best of our knowledge, this is the first work to understand and measure the commonsense reasoning capability in neural machine translation. The contributions of this paper can be summarized as follows:
\begin{itemize}
\setlength{\itemsep}{0pt}
\setlength{\parsep}{0pt}
\setlength{\parskip}{0pt}
\item We build a test suite\footnote{The built commonsense test suite will be publicly available at https://github.com/tjunlp-lab/CommonMT.} to examine the ability of neural machine translation in commonsense reasoning, which provides a benchmark testbed for tracking progress in this direction.
\item Based on our experiments and analyses on evaluating commonsense reasoning in NMT, we find that: 1) commonsense reasoning related to lexical ambiguity and contextual syntactic ambiguity is more difficult than contextless syntactic ambiguity; 2) although the commonsense reasoning accuracy is higher than 50\%, the reasoning consistency rate is far lower than 50\% (random guess).
\end{itemize}

\section{Related work}
\noindent
We briefly review recent efforts related to commonsense reasoning in NLP. We refer readers to \citet{survey-commonsense}'s article for a thorough survey in this area.

\noindent
\textbf{Commonsense Datasets}
%a general understanding of how the physical world works (i.e., intuitive physics); a basic understanding of human 
%motives and behaviors (i.e., intuitive psychology); and knowledge of the common facts that an average 
%adult possesses这是这篇文章的原文

\noindent
According to ~\citet{DBLP:journals/corr/abs-1810-07528}, commonsense knowledge normally consists of a general theory of how the physical world works and a basic understanding of human motives and behaviors. In recent years, a wide variety of datasets on the two kinds of commonsense knowledge have been proposed. 
~\citet{sap-etal-2019-social} introduce Social IQA, containing 38k multiple choice questions for probing the commonsense reasoning about emotional and social in people's daily life. Similarly, Event2mind and Atomic ~\citep{event2mind,ATOMICAA} focus on inferred knowledge in the form of {\em if-then} to reason about people's daily life behavior. For datasets on physical common sense, PIQA ~\citep{PIQARA} on commonsense phenomena in the physical world contains 21K QA pairs. SWAG and HellaSwag ~\citep{swag,hellaswag} are datasets on commonsense NLI, where materials from video subtitles and wikihow articles are used to construct cloze tests. ~\citet{AbductiveCR} propose a dataset for abductive reasoning on events. The well-known Winograd Schema Challenge (WSC) test set~\citep{Levesque:2012:WSC:3031843.3031909,DBLP:journals/corr/abs-1907-10641} focus on solving the commonsense problems in the form of coreference resolution. Different from them  on monolingual data, we provide a bilingual commonsense test suite for machine translation.

\noindent
\textbf{Commonsense Reasoning in NLP}

\noindent
In addition to common sense datasets, we have also witnessed that commonsense knowledge has been recently explored in different NLP tasks. Just to name a few,~\citet{DBLP:journals/corr/abs-1806-02847},~\citet{he-etal-2019-hybrid} and ~\citet{attention-is-not-need}  use language models trained on huge text corpora to do inference on the WSC dataset. ~\citet{ding-etal-2019-event-representation} use commonsense knowledge in Atomic ~\citep{ATOMICAA} and Event2mind ~\citep{event2mind} on downstream tasks such as script event prediction. ~\citet{bi-etal-2019-incorporating} exploit external commonsense  knowledge from ConceptNet \citep{Speer2016ConceptNet5A}) in machine reading comprehension.

\noindent
\textbf{Commonsense Reasoning Evaluation}

\noindent
With pre-trained language models, like BERT ~\citep{bert}, GPT-2 ~\citep{gpt2} being widely used in various NLP tasks, studies have been performed to examine the commonsense reasoning capability in pre-trained neural language models. ~\citet{2019-make} and ~\citet{Zhou2019EvaluatingCI} propose to measure the success rate of the pre-trained language models in commonsense inference by calculating LM probabilities.  Two sentences which are used to test commonsense inference differ only in commonsense concepts. ~\citet{Feldman2019CommonsenseKM} further explore unsupervised methods to generate commonsense knowledge using the world knowledge of pre-trained language models. Our commonsense reasoning evaluation resonates with these evaluation efforts.

\noindent
\textbf{Commonsense Knowledge and Reasoning in Machine Translation}

\noindent
Commonsense knowledge has long been acknowledged as an indispensable knowledge source for disambiguation in machine translation ~\citep{1960hill,DBLP:journals/cacm/DavisM15}. Knowledge-based machine translation (KBMT), one of the popular machine translation paradigms in 1980s, lays much stress on extra-linguistic world knowledge in machine translation ~\citep{1989-knowledge-translation}. Large ontology that is constructed either manually or automatically to provide world knowledge is one of essential components in KBMT ~\citep{knight1994building}. 

As data-driven machine translation, such as statistical machine translation (SMT) and neural machine translation,  becomes de facto standard in machine translation, world knowledge has been less explicitly explored. Only a few studies have indirectly and partially exploited world knowledge in SMT or NMT, by incorporating linked open data resources such as DBpedia and BabelNet into SMT with modest improvements ~\citep{2016-using,2017ImprovingMT,2018MachineTU}.  In addition to world knowledge, discourse-level context has also been leveraged to enhance translation coherence and consistency. For example, \cite{long-etal-2020-shallow,he-etal-2022-evaluating} demonstrate that explicitly modeling discourse relations can benefit MT. \cite{long-etal-2020-ted} developed a Chinese discourse corpus (TED-CDB) of TED talks, revealing that genre-specific discourse differences pose challenges to translation. \cite{long-webber-2022-facilitating,long-etal-2024-multi,long2024leveraginghierarchicalprototypesverbalizer} enabling improvement of cross-sentence references and encouraging the integration of discourse knowledge into MT.

\section{Commonsense Reasoning Test Suite for Machine Translation}
In this section, we discuss the design and construction of the test suite, including the rules and steps for building this test suite.
\setlength{\lineskip}{0em} 

\subsection{Test Suite Design}
\label{test-suite-design}
Different from commonsense reasoning in Winogram Schema Challenge ~\citep{Levesque:2012:WSC:3031843.3031909} or sentence reasonability judgment (i.e., ``He put a turkey
into the fridge” vs. ``He put an elephant into the fridge”)
~\cite{2019-make}, where commonsense reasoning normally happens in one language, commonsense reasoning in NMT can be done either in the encoding of the source language (i.e., encoding reasonable source representations) or in the decoding of the target language (i.e., producing reasonable target outputs). As it is difficult to detect whether reasonable senses are identified and encoded in the encoder, we check target outputs from the decoder to test the commonsense reasoning capability of NMT. This is the first rule that we follow to design the test suite. 

In the second rule for building the test suite, we manually create source sentences with ambiguity that requires commonsense reasoning. Inspired by ~\citet{schwartz} and ~\citet{2012IntegrationOW}, we ground the commonsense reasoning test on two types of ambiguity: lexical and syntactic ambiguity (LA and SA), which are common in machine translation. An example in LA is the ``batter'' in ``she put the batter in the refrigerator” (food material vs. baseball player). SA relates to structures, for instance, ``I saw a man swimming on the bridge” (I was standing on the bridge vs. The man was swimming on the bridge). We further refine SA into contextless (e.g., Example (2) in Figure \ref{figure-introduction}) and contextual SA (e.g., Example (3) in Figure \ref{figure-introduction}). The former can be correctly interpreted by resorting to commonsense knowledge while the latter cannot be interpreted uniquely if no more context is given.

\begin{figure}[t]
\setlength{\belowcaptionskip}{-0.6cm} 
\centering
\includegraphics[scale=0.65]{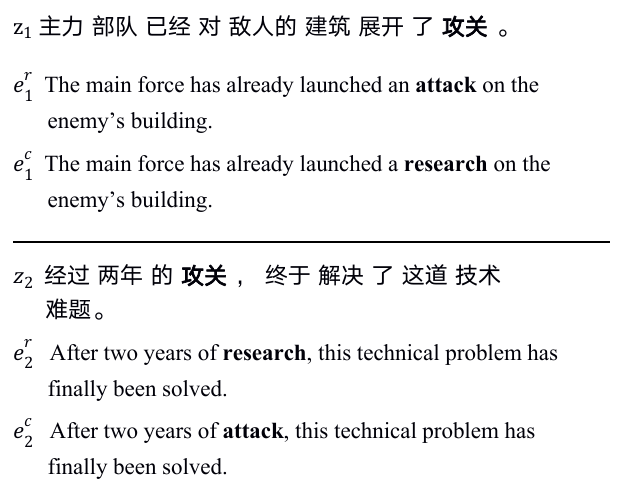}
\caption
{An example block in the LA test set.}
\label{figure-1}
\end{figure}

\begin{figure}[t]
\setlength{\belowcaptionskip}{-0.5cm} 
\centering
\includegraphics[scale=0.65]{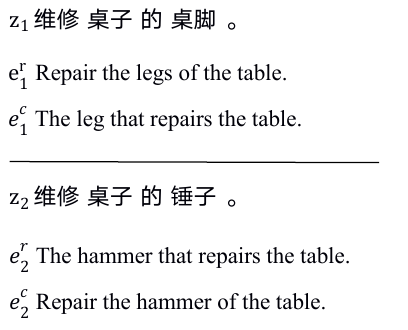}
\caption
{An example block in the contextless SA test set.}
\label{figure-3}
\end{figure}

\begin{figure}[t]
\setlength{\belowcaptionskip}{-0.6cm} 
\centering
\includegraphics[scale=0.65]{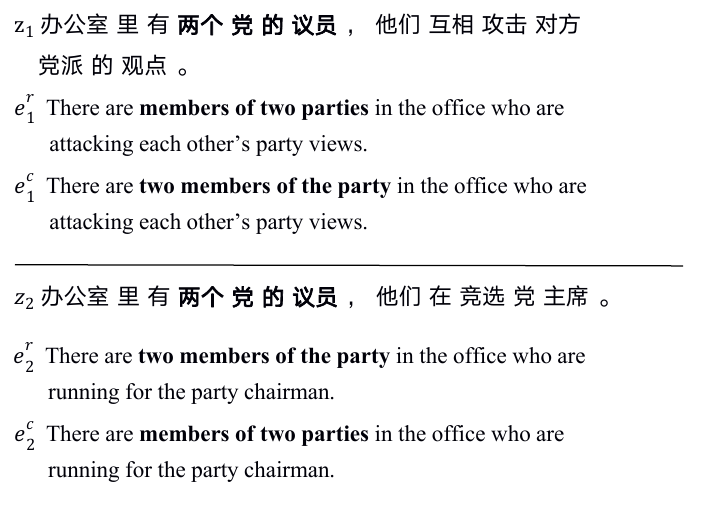}
\caption
{An example block in the contextual SA test set.}
\label{figure-2}
\end{figure}

The third rule that we conform to is to 1) create two contrastive source sentences for each lexical or syntactic ambiguity point, where each source sentence corresponds to one reasonable interpretation of the ambiguity point, and 2) to provide two contrastive translations for each created source sentence. This is similar to other linguistic evaluation by contrastive examples in the MT literature \citep{etal-2019-linguistic,evaluate-discourse-pheonmena,etal-2018-large,2017-grammatical}. These two contrastive translations have similar wordings: one is correct and the other is not correct in that it translates the ambiguity part into the corresponding translation of the contrastive source sentence. This translation makes sense in the contrastive sentence but not in the sentence in question. Examples of contrastive source sentences and contrastive translations for each source sentence are shown in Figure \ref{figure-1}, \ref{figure-3} and \ref{figure-2}. 
 
Finally, we have hired two linguistic experts to construct ambiguous source sentences and two professional human translators to provide contrastive translations for each source sentence. We ask them to create and translate with diverse words as much as possible and hire an extra linguistic expert and translator to review and double check source sentences and target translations after the two experts and translators cross check with each other.   
%感觉好多地方都搬了人家的说法和用词了--problem
%双引号标点全错了..后面再改吧..-problem
%这个表的字体有点问题---problem
%表的空白过大--problem

\noindent
\subsection{Lexical Ambiguity Test Set} 
To construct this test set, we select words from a Chinese polysemous dictionary\footnote{\href{https://drive.google.com/file/d/1Ejs9xUjHMKTUd8h9W4chALwCox45uGP1/view?usp=sharing}{Download link for the Chinese polysemous dictionary}} so that the selected words have multiple interpretations. We avoid selecting words that are semantically close to one another in order to maintain diversity of the test set. We do not select words that are polysemous in Chinese but translated into the same words in English. Words that are translated into very different English words in different context and require commonsense knowledge to disambiguate are preferred. 

This test set contains 200 example blocks. Each block is composed of two contrastive triples ($z_1$, $e^r_1$, $e^c_1$) and ($z_2$, $e^r_2$, $e^c_2$). As shown in Figure \ref{figure-1}, $z_1$ and $z_2$ are contrastive with each other as they contain the same ambiguous word with different meanings. $e^r_.$ and $e^c_.$ are contrastive translations where the former is correct while the latter not. $e^c_1$ and $e^c_2$ are wrong translations in that they incorrectly interpret the ambiguous word in the way of $e^r_2$ and $e^r_1$ respectively. A selected polysemous word is used in only one example block.
%像figure3那种图形宽度是没法调整的吗？
%还有检查引用列表，好像有不该混进来的东西...--problem

\noindent
\subsection{Syntactic Ambiguity Test Sets}
As mentioned before, we have two types of test sets for syntactic ambiguity: contextless and contextual SA. Before we construct the two test sets, we select Chinese structures that are typically ambiguous, just like PP attachment in English (e.g., ``He ate the apple in the refrigerator” from \citet{schwartz}).

\citet{14-jiegouqiyi} has deeply investigated syntactic ambiguity in Chinese and has found 26 structures that tend to generate sentences with different interpretations, such as ``noun phrase + de (a Chinese particle) + shi (is) + noun phrase". From them, we use 12 structures to construct contrastive examples, where the subtle differences in Chinese can be clearly detected in English after translation.

%For instance, one example of SA in Chinese is \emph{``发明的是一个机器人''}. Humans can judge based on commonsense that robots do not have the ability to invent. Generally, robots are invented, so the corresponding translation for this sentence is \emph{``A robot was invented''} rather than \emph{``It is a robot that invents it''}. This syntax can be summarized as:N +的+是+N (See Appendix (§A) for all the 12  syntax). 

With these 12 structure templates with potential syntactic ambiguity, we manually create 225 example blocks for the contextless SA test set and 175 blocks for the contextual SA test set. Examples of these two test sets are listed in Figure \ref{figure-3} and \ref{figure-2}. Similar to the LA test set, each block is composed of two contrastive triples where two translations for each source sentence are also contrastive with each other in the way that we translate sentences in the LA test set. For the blocks in the contextless test set, we make sure that each ambiguous source sentence can be correctly interpreted with commonsense knowledge. We do not need extra context information for disambiguation. In contrast, we have to resort to additional context to interpret sentences in the contextual SA test set. 
\begin{table*}[!t]
\flushleft
%\smal
%\begin{tabular}{cp{2em}p{6em}p{7em}p{8em}}
\begin{tabular}{l|llll}
\toprule %添加表格头部粗线
Test set & \#triples & \#unique tokens & Average tokens per sentence & total token numbers\\
\hline
LA& 400& 1,246/1,139/1,140 &7.3/9.1/9.1 & 2,920/3,640/3,640 \\
CL-SA & 450 & 838/738/741 &5.2/6.3/6.3&2,340/2,835/2,835 \\
CT-SA & 350 & 1,083/997/997 &11.1/13.5/13.5&3,885/4,725/4,725 \\
TOTAL& 1,200 & 2,570/2,050/2,063&7.6/9.3/9.3&9,120/11,160/11,160\\
\toprule %添加表格头部粗线
\end{tabular}
\caption
{Statistics on the test suite. Numbers a/b/c denote the corresponding number in source sentences/correct translations/incorrect translations. LA: lexical ambiguity; CL-SA: contextless SA; CT-SA: contextual SA.}
\label{table-2}
\end{table*}

\begin{table*}[!htb]
\setlength{\belowcaptionskip}{-0.5cm} 
\centering
\scriptsize
%\footnotesize
\begin{tabular}{l|l|p{26em}|l}
Category & Descriptions & Examples&\% \\
\hline
Properties&properties of objects&你/you 嘴/mouth 太快了/too fast&25.9\\

Behaviors&Behaviors that objects will take in a particular situation&鸡/ chicken 不/not 吃了/eat 因为/because 这只鸡/the chicken 已经/had already 吃了/eat 太多了/too much.&25.2\\

Taxonomy&Systematic classification of objects and concepts&今年/this year 风调雨顺/weather is good 农民的秋景/the harvest of the farmers' autumn 一定/must be 很好/very good.&21.1 \\

Action&Some actions an object may be involved in&健康的/ healthy 医生/doctor 正在/is doing 手术/surgery.&15.8 \\

Structures&Object A is part of Object B&削/Cut 西瓜的/the watermelon 皮/skin.&8.1 \\

Emotions&Description of people's psychological activities and emotions&她/she 留下/leave 眼泪/tears 倾倒/pour out 她的/her 委屈/grievances.&2.6 \\

Procedural& The type of common sense exercised in the performance of a task &学生/students 被调查/were investigated 因为/because 这些学生/these students 是/were 这个事件的/the incident 目击者/witnesses.&1.3  \\

\end{tabular}
\caption
{Commonsense knowledge categories and their percentages in the test sets.}
\label{table-3}
\end{table*}

\section{Test Suite Analysis}

We provide statistical analyses on the built test suite, which cover its size, distribution of knowledge types and the reasoning accuracy of pretrained language models on target translations of target translations of this test suite.
\noindent
\subsection{General Statistics}
Statistics on the built test suite are displayed in Table \ref{table-2}. We show the number of triples, the number of unique tokens, and the average number of tokens per sentence in each test set. Although sentences in the test suite are not very long, they are very challenging to be correctly translated as commonsense reasoning is involved, which will be verified in our experiments. 

% \begin{table}
% \centering
% %\small
% \begin{tabular}{ll}
% \toprule  %添加表格头部粗线
% POS tag &  Percentage \\
% \hline
% VERB&59\% \\
% NOUN&30\% \\
% ADJ&11\% \\
% \toprule  %添加表格头部粗线
% \end{tabular}
% \caption
% {POS tag distribution of ambiguous words in the LA test set. }
% \label{table-1}
% \end{table}

% \subsection{Lexical Ambiguity Word Type}
% We perform part-of-speech (POS) tagging on Chinese sentences in the LA test. After that, we calculate the percentages of POS tags of ambiguous words, which are shown in Table \ref{table-1}.  It is worth noting that the POS tag of the ambiguous word in each block are the same across the two triples in the block. 

\subsection{Commonsense Knowledge Type}

\citet{DBLP:journals/sigmod/TandonVM17} categorize commonsense knowledge into different types. Following their taxonomy of commonsense types, we compute the percentage of each type in our test suite, as shown in Table \ref{table-3}. Commonsense knowledge on properties,  behaviors and taxonomy of objects/concepts are the top 3 commonsense knowledge types involved in our test sets.  %这个地方还存在疑问,因为文章里面没有说到是7种% and also refer to their demonstration at CIKM2017\footnote{http://people.mpi-inf.mpg.de/~ntandon/presentations/cikm-2017-tutorial-commonsense/commonsense.pdf}. Table \ref{table-3} presents the commonsense categories we used,their definition,an example and their frequency in the test sets. 
%For more details,please refer to the Appendix ~\ref{appendix-a}.
%这个附录怎么加..
%那个实验的部分也放到附录讲吧？但不知道要不要提个problem的问题..
%这个论文怎么引入啊..不知道.
%那个鲁棒性的实验怎么补加很好奇..

\begin{table}
\setlength{\belowcaptionskip}{-0.5cm} 
\centering
%\scriptsize
\small
%\begin{tabular}{c|p{4em}p{4em}p{4em}|c}
\begin{tabular}{l|lll|l}
\bottomrule
 &LA & CL-SA& CT-SA&Total \\
\hline
Random&0.500&0.500&0.500&0.500\\
GPT&0.775&0.650&0.597&0.678 \\
GPT-2 base&0.803&0.642&0.606&0.688 \\
GPT-2 medium&0.798&0.648&0.611&0.690 \\
BERT-base&0.788&0.642&0.611&0.684\\
BERT-large&\textbf{0.818}&\textbf{0.682}&\textbf{0.623}&\textbf{0.712} \\
\toprule

\end{tabular}
\caption
{Commonsense Reasoning accuracy of pretrained language models on the 1,124 instances of the test suite.}
\label{table-plm}
\end{table}

\subsection{Evaluation of Pretrained Language Models on the Test Suite}

In our test suite, we find that target translations of 93.7\% instances (1,124 of 1200 test instances) can be determined if they are correct only from translations themselves (i.e., by performing commonsense reasoning), without reference to the corresponding source sentences. This is exactly what we want the test suite to be like as the purpose of this test suite is to evaluate commonsense reasoning rather than the ability of NMT in exploring source context for translation disambiguation not related to common sense. This is also consistent with the first rule for building the test suite: evaluating commonsense reasoning from the target side. 
Since the reasonability of these translations can be determined only from themselves, we want to know how challenging they are for pretrained language models in terms of commonsense reasoning. Hence, we evaluate state-of-the-art language models pretrained on large-scale data, including BERT \citep{bert}, GPT \citep{gpt}, and GPT-2 \citep{gpt2}, on these 1,124 translation pairs (pairs of reference and contrastive translations). For notational convenience, we still use the test suite to refer to these instances as only 76 cases are excluded for this evaluation.

Following \citet{2019-make} and \citet{Zhou2019EvaluatingCI}, for each pair $(e^r$, $e^c)$, we use a pretrained language model to compute the language model score of the two translations.
The translation with a higher score is labelled as the correct one by the language model.
By comparing these labels with ground-truth labels, we can obtain the commonsense reasoning accuracy of the corresponding language model on these instances.
%We also conduct human evaluation on the dataset. In each example, two subjects judge whether a translation is more reasonable than the other without seeing the corresponding source sentence. If and only if when the two subjects independently judge that a translation is correct from the perspective of commonsense reasoning, we treat it as a positive translation. In this way, we can calculate the accuracy of human on these instances. 

Results are shown in Table \ref{table-plm}. All language models are better than random guess, validating the commonsense reasoning ability of them. They perform worse on the contextual SA test than on the other two test sets, demonstrating the difficulty in cross-sentence commonsense reasoning. BERT-large achieves the highest accuracy, 0.712. The number of parameters of BERT-large is equal to that of GPT2-medium, almost 3 times as large as that of GPT-2 base and BERT-base (340M vs. 117M). We conjecture that the reason for the superiority of BERT models over GPT/GPT-2 models is due to bidirectional context in BERT, which resonates with the findings of \citet{Zhou2019EvaluatingCI}. The accuracies of all pretrained language models are all lower than 72\%. This suggests that our test suite  is very challenging in commonsense reasoning even for language models trained on an amount of data.
%The human performance is lower on the contextless SA test set than that on the contextual SA test set. The reason is .....
\label{e-prelm-model}

\section{Experiments}

In this section, we conducted extensive experiments to evaluate the commonsense reasoning capability of state-of-the-art neural machine translation on the built test suite. 
%\subsection{Training Data for MT Models}

\subsection{Experimental setup}

%这部分涛哥写吧..
We adopted the CWMT Chinese-English corpus\footnote{Available at: \href{http://nlp.nju.edu.cn/cwmt-wmt}{http://nlp.nju.edu.cn/cwmt-wmt}} of news domain as training data for NMT systems.
This corpus contains 9M parallel sentences. We used byte pair encoding compression algorithm (BPE) \cite{sennrich2016neural} to process all these data and restricted merge operations to a maximum of 30k.

We trained two neural machine translation models on the training data: RNNSearch \cite{DBLP:journals/corr/BahdanauCB14} and Transformer \cite{vaswani2017attention}. 

We used the Transformer base model with 6 layers and 8 self-attention heads per layer.
As for RNNSearch, we employed neural architecture with 4 layers of LSTM and 512-dimension hidden states.
We used Adam \cite{DBLP:journals/corr/KingmaB14} to train both NMT models. ${\beta} 1$ and ${\beta} 2$ of Adam were set to 0.9 and 0.999, the learning rate was set to 0.0005, and gradient norm was set to 5. To take full advantage of GPUs, we batched sentences of similar lengths. We trained both models on a single machine with 8 1080Ti cards. Each mini-batch contained 32,000 tokens.
During decoding, we employed the beam search algorithm and set the beam size to 5.

\subsection{Evaluation Metrics}

For translation performance evaluation, we used sacrebleu \cite{post-2018-call} to calculate case-sensitive BLEU-4 \cite{Papineni2001BleuAM}. 
%For commonsense reasoning accuracy evaluation in NMT generation, we forced the NMT models to output probability for each translation in each contrastive translation pair. Models are tested as to whether they assign a higher probability to the reference translation than to the contrastive example. This method is convenient for automatic evaluation. Therefore, we use this method in experiments for comparative evaluation of future work.

To evaluate the commonsense reasoning accuracy of NMT on the test suite, we applied NMT models to score each pair $(s,t)$ as follows:
\begin{equation}\label{eual}
Score(t|s)=\frac{1}{|t|}\sum_{i=0}^{|t|} {\rm log}p(t_i|t_{<i},s)
\end{equation}
where $p(t_i|t_{<i},s)$ is the probabilty of the target word $t_i$ given the target history and source sentence.
Given a triple $(z, e^r, e^c)$, if an NMT model scores the reference translation higher than the contrastive translation (i.e., $Score(e^r|z) > Score(e^c|z)$), the NMT model is believed to make a correct commonsense reasoning prediction. This is reasonable as $e^r$ and $e^c$ are only different in words or structures related to the lexical or syntactical commonsense ambiguity point as described in Section \ref{test-suite-design}. By scoring each triple with an NMT model, we can measure the commonsense reasoning accuracy of the model on our test suite.

\subsection{Results}

BLEU scores for the two NMT models are given in Table \ref{table-4}. Commonsense reasoning results on the test suite are provided in Table \ref{table-5.2}.

% \begin{table}
% \centering

% \begin{tabular}{ccccc}
% \toprule

% % & \multicolumn{4}{c}{Accuracy(\%)}\\
% % Model&1&2&3&4\\
% % \midrule
% % LSTM&54&56.44&53.43&54.75\\
% % Transformer&56.75&65.33&55.43&59.58\\
% % Google Translate&57.75&72.22&54.29&62.17\\
% % \bottomrule
% % \end{tabular}
% % \caption
% % {Accuracy for the test sets. 1/2/3/4 represents represents Represents Lexical ambiguity, Syntax without context,Syntax with context and the total accuracy score.}
% % \label{table-5}
% % \end{table}

%还有SMT的结果跟分析没有添加

% \begin{figure}[t]
% \centering
% \includegraphics[scale=0.45]{images/figure_result_1.eps}
% \caption
% {Commonsense reasoning accuracy on the test sets.}
% \label{figure-8}
% \end{figure}

From the table and figure, we can observe that
\begin{itemize}
\setlength{\itemsep}{0pt}
\setlength{\parsep}{0pt}
\setlength{\parskip}{0pt}
\item Both BLEU and commonsense reasoning accuracy clearly show that Transformer is better than RNNSearch.

\item Both RNNSearch and Transformer perform better on the contextless SA than on the contextual SA according to the commonsense reasoning accuracy. This is consistent with the results of pretrained language models shown in Table \ref{table-plm}, suggesting that cross-sentence commonsense reasoning is also challenging for NMT. Notice that the commonsense reasoning accuracy of pretrained language models cannot be directly compared to that of NMT models due to different evaluation procedure, mechanisms for commonsense reasoning and different test data. The BLEU scores on the contextless SA test set are lower than those on the contextual SA. We conjecture that this is because the contextless SA test set consists of very short sentences. Wrongly translated words therefore have a very big impact on BLEU scores.
\begin{table}
\centering
\small
\begin{tabular}{lllll}
\toprule
& LA & CL-SA & CT-SA &Total\\
\midrule
RNNSearch &25.82&21.59&27.98&25.86 \\
Transformer &31.97&27.84&31.30&30.75 \\
\bottomrule
\end{tabular}
\caption
{BLEU scores on the test sets.}
\label{table-4}
\end{table}
\begin{table}
\setlength{\belowcaptionskip}{-0.5cm} 
\centering
\small
\begin{tabular}{lllll}
\toprule
& LA & CL-SA & CT-SA& Total \\
\midrule
RNNSearch &0.543&0.569&0.551&0.555 \\
Transformer &0.565&0.656&0.571&0.601 \\
\bottomrule
\end{tabular}
\caption
{Commonsense Reasoning accuracy on the test sets.}
\label{table-5.2}
\end{table}
\item The performance gap between Transformer and RNNSearch on the CL-SA test set is larger than that on the other two test sets. The reason might be that the self-attention mechanism allows Transformer to more easily detect collocations (e.g., ``leg" and ``table" in Figure \ref{figure-3}) for disambiguation on the CL-SA test set. Many CL-SA cases can be disambiguated by collocations according to our observation on this test set.

\item Compared with the relative BLEU improvement of Transformer over RNNSearch, the relative improvement in terms of commonsense reasoning accuracy is smaller (8.2\% vs. 18.91\% in BLEU), indicating that more efforts are expected to not only improve translation quality in terms of BLEU but also to enhance commonsense reasoning ability in NMT. 
\end{itemize}

\begin{figure}[t]
\setlength{\belowcaptionskip}{-0.5cm} 
\centering
\includegraphics[scale=0.3]{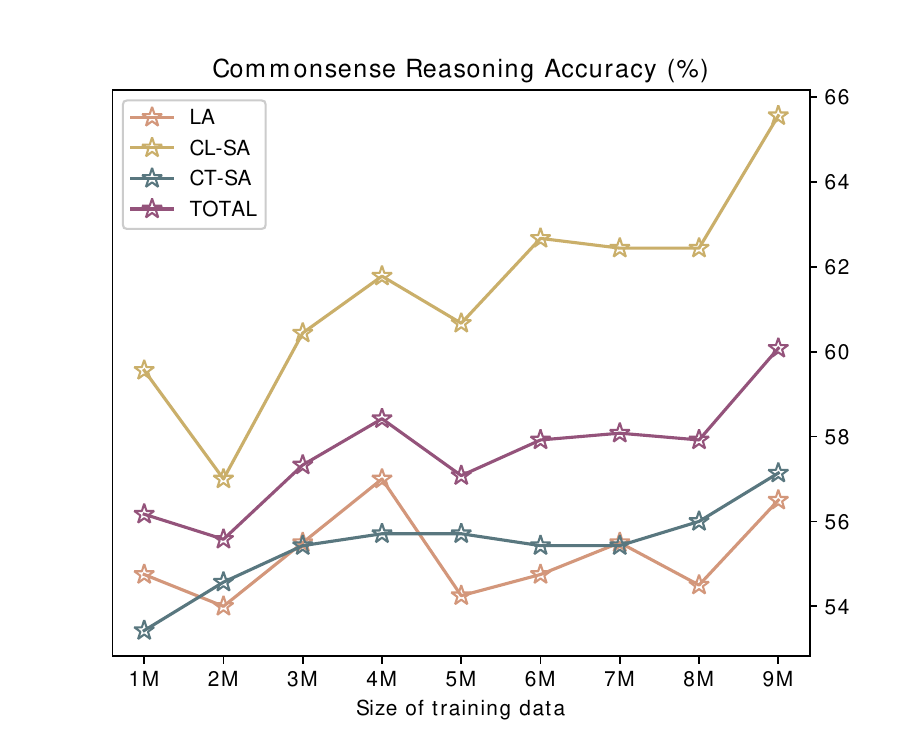}
\caption
{Commonsense Reasoning accuracy of the Transformer on the test sets with different size of training data.}
\label{figure-4}
\end{figure}

%这里有些东西没有找到
\subsection{Effect of the Size of Training Data}

We conducted experiments to investigate the impact of the amount of training data on the commonsense reasoning performance of the state-of-the-art NMT model Transformer. Results are displayed in Figure \ref{figure-4}. Generally, with the increase of training data, The common-sense reasoning ability of NMT systems rises too. 
Although we used all CWMT Chinese-English training data to train NMT, we didn't have a chance to see that the commonsense reasoning accuracy tends to level off. We conjecture that the growth has the potential to continue. We leave using more data to measure the growth momentum of NMT commonsense reasoning to our future work. 

Yet another finding from Figure \ref{figure-4} is that the commonsense reasoning performance on the contextless SA test set is always higher that the other two test sets. As shown in the last subsection, the reasons for this may be due to shorter sentences and collocations in this test set. 
\label{sect-1} 

\subsection{Effect of Sentence Length}

\begin{figure}[t]
\setlength{\belowcaptionskip}{-0.5cm} 
\centering
\includegraphics[scale=0.35]{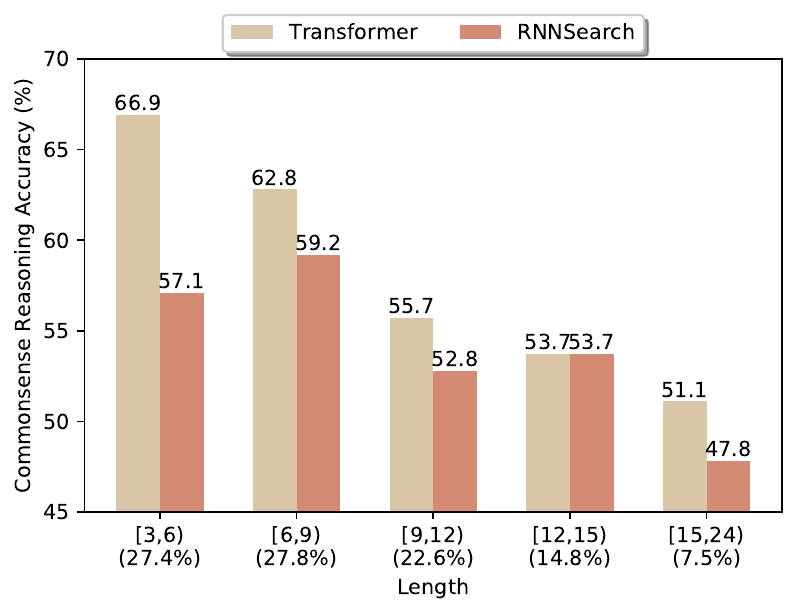}
\caption
{Commonsense Reasoning accuracy against the length of source sentences. The percentage of each group is shown under the corresponding length interval. }
\label{figure-distance}
\end{figure}

We carried out an analysis on the impact of the length of source sentences on commonsense reasoning. We divided the test suite into 5 groups according to the length of source sentences. The results are shown in Figure \ref{figure-distance}. Generally, Transformer is better than RNNSearch in almost all length groups. With the length of source sentences increasing, the commonsense reasoning performance tends to go down. This may suggest that long-distance or cross-sentence commonsense reasoning is more challenging for NMT than short-distance reasoning, which is consistent with our finding on the CL-SA test set.

% \begin{figure}[t]
% \setlength{\belowcaptionskip}{-0.5cm} 
% \centering
% \includegraphics[scale=0.35]{images/figure_result_2.eps}
% \caption
% {Commonsense Reasoning accuracy of the Transformer on the different commonsense knowledge types.}
% \label{figure-5}
% \end{figure} 
% \begin{table}
% \setlength{\belowcaptionskip}{-0.5cm}
% \centering
% \small
% \begin{tabular}{lllll}
% \toprule
% & LA & CL-SA & CT-SA&Total \\
% \midrule
% RNNSearch &0.24&0.31&0.27&0.27 \\
% Transformer &0.26&0.39&0.27&0.31 \\
% \bottomrule
% \end{tabular}
% \caption
% {Rates of Reasoning consistency on the three test sets. }
% \label{table-5.6}
% \end{table}

\begin{figure*}[tb] 
\setlength{\belowcaptionskip}{-0.2cm} 
  \begin{minipage}{0.5\textwidth} 
    \centering 
    \includegraphics[width=0.8\textwidth]{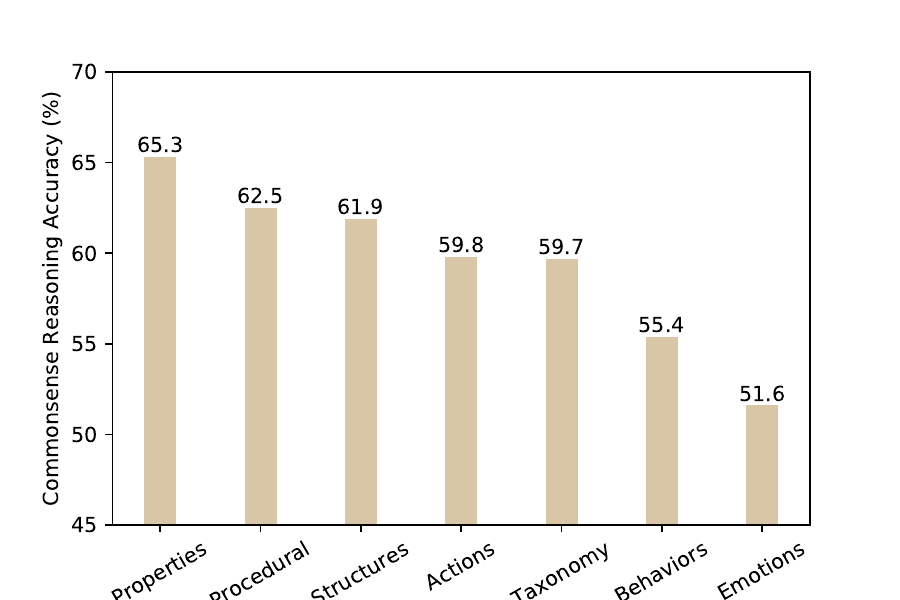} 
    \caption{Commonsense Reasoning accuracy of the \\ Transformer on the different commonsense knowle-\\dge types.} 
    \label{figure-5} 
  \end{minipage}% 
\makeatletter\def\@captype{table}\makeatother
  \begin{minipage}{0.55\textwidth}
    \centering
    \begin{tabular}{lcc}
\toprule
& RNNSearch & Transformer \\
\midrule
LA&0.24&0.26\\
CL-SA &0.31&0.39 \\
CT-SA &0.27&0.27\\
Total&0.27&0.31\\
\bottomrule
\end{tabular}
\newline
\caption
{Rates of Reasoning consistency on the \\ \hspace{2em} three test sets.}
\label{table-5.6}
  \end{minipage} 
\end{figure*}

% \begin{table*}
% \setlength{\belowcaptionskip}{-0.65cm} 
% \centering
% \footnotesize
% \begin{tabular}{lll}
% \toprule
% Error type&Example&\% \\
% \midrule
% Common sense errors&Origin: 公园里有\underline{三个幼儿园的孩子}，总共有6个孩子在做游戏。\\
% &Reference: There are children of three kindergartens in the park, and \\a total of six children are playing games.&\\
% &Transformer: There are three kindergarten children in the park, a total \\of 6 children are playing games.&\\
% \hline
% Ordinary meaning errors&Origin: 这个\underline{工程}已经下马。 
% &22.7\% \\
% &Reference: This project has been abandoned.&\\
% &Transformer: The \underline{factory} is already off.&\\
% \hline
% other errors&Origin: 放学后这几个学生约好一起去水库\underline{下水}。&5.7\%\\
% &Reference: After school, these students agreed to go to the reservoir together \underline{for swimming}.&\\
% &Transformer: After school the students made an appointment to go to the reservoir. &\\
% \bottomrule
% \end{tabular}
% \caption
% {Translation error types. Words related to translation errors are underlined. }
% \label{table-5.9}
% \end{table*}

\begin{table*}[]
%\small
\resizebox{\linewidth}{!}{
\begin{tabular}{lll}
\toprule
Error type &
  Example &
  \multicolumn{1}{c}{\%} \\ \hline
Common sense errors &
  \begin{tabular}[c]{@{}l@{}}Origin: 公园里有\underline{三个幼儿园的孩子}，总共有6个孩子在做游戏。\\ Reference: There are children of three kindergartens in the park, and a total of six children are playing games.\\ Transformer: There are \underline{three kindergarten children} in the park, a total of 6 children are playing games.\end{tabular} &
  71.6\% \\ \hline
Ordinary meaning errors &
  \begin{tabular}[c]{@{}l@{}}Origin: 这个\underline{工程}已经下马。\\ Reference: This project has been abandoned.\\ Transformer: The \underline{factory} is already off.\end{tabular} &
  22.7\% \\ \hline
Other errors &
  \begin{tabular}[c]{@{}l@{}}Origin: 我写了六天字帖。\\ Reference: I wrote copybooks for six days.\\ Transformer: I wrote six days.\end{tabular} &
  5.7\% \\
 \bottomrule
\end{tabular}}
\caption {Translation error types. Words related to translation errors are underlined.}
\label{table-5.9}
\end{table*}

\subsection{Effect of Commonsense Knowledge Types}

%Like humans, when the commonsense knowledge involved is more complicated, the performance of the model will decrease. We have analyzed and verified this conjecture, and we have measured the accuracy of the model for classification of commonsense types, as shown in Fig.\ref{figure-5}. In judging the properties of this object and whether an object belongs to another object, the accuracy rate is the highest two, which is consistent with human intuition, because such data is often directly reflected in a sentence, and then which follows is commonsense of action. The correct rate of actions that may be taken by a thing is higher than that of behavior, because behavior needs to judge the action that a thing will take under certain circumstances.
Finally, we analyzed the commonsense reasoning capability of Transformer on different commonsense knowledge types. Studying different types of common sense can help us understand what kind of commonsense knowledge is more needed to solve commonsense reasoning problems in NMT. The results are shown in Figure \ref{figure-5}. Transformer-based NMT obtains relatively good results on commonsense reasoning on properties, structures, actions, but performs badly on reasoning on behaviors and emotions. 
%For NMT models, translation that require Emotion knowledge are the most challenging ones. One possible explanation is that the inference over emotion knowledge reflects the inner activities of people in the interaction with the world, which is challenging for MT models to build the connection without enough training corpus for emotion knowledge specifically;

%As for the taxonomy knowledge about things, when we construct the test set, we have commonsense of classification, because we need to judge whether an thing belongs to or does not belong to a certain category. This requires more comprehensive knowledge, rather than it is limited to the definition in the general classification sense ~\citep{Ponzetto2011TaxonomyIB}, so the accuracy rate is not high, and the emotional emtion and procedural categories of people do not have high confidence because the data is less than the others,so we did not add this part of the data.
%This shows that compared with humans who can easily reason about commonsense, but for machine translation models, some subtle commonsense knowledge is challenging.
%先随便取个标题了...
% \begin{figure*}[t]
% \centering
% \scriptsize
% \includegraphics[scale=0.45]{images/analyse.eps}
% \caption
% {Examples of different errors.}
% \label{figure-6}
% \end{figure*}

\section{Further Analysis}

\setlength{\abovedisplayskip}{-5pt}
\subsection{Analysis on Reasoning Consistency }

Our test suite contains 600 example blocks, each of which focuses on only one LA/SA ambiguity point. For the two reasonable interpretations $(z_1, z_2)$ of a given ambiguity point, NMT models need to make two translation predictions: one for $(e_1^r, e_1^c)$ and the other for $(e_2^r, e_2^c)$. If they choose $e_1^r$ and $e_2^r$ (both right reasoning predictions) or $e_1^c$ and $e_2^c$ (both wrong reasoning predictions), we treat this as a consistent reasoning, otherwise inconsistent. Partially inspired by \citet{Zhou2019EvaluatingCI}, we conducted an analysis on reasoning consistency. 

We counted the times that a tested NMT model made consistent reasoning predictions and calculated the consistency rate on the three test sets. Results are shown in Table \ref{table-5.6}. Disappointingly, the reasoning consistency rates for both RNNSearch and Transformer are lower than random guess (0.5). On the contextless SA test set where both NMT models have higher reasoning accuracies, the rates of reasoning consistency are also higher than those of the other two test sets. 

% The results are shown in Table \ref{table-5.6}. In theory, a model equipped with relevant commonsense should give consistent predictions on a pair of test case. However, we find that both of the models reach low consistency. In fact, their consistency is below the random baselines. We believe that this reveals that the NMT model actually lacks an understanding of commonsense. For the same pattern of vocabulary and stntax, it tends to prefer the same translation without considering the actual situation. when the sentence length is longer and the situation is more complicated. The accuracy on CT-SA is lower than CL-SA also confirms this.

\subsection{Analysis on Translation Errors}

We have already automatically evaluated commonsense reasoning in NMT with both reasoning accuracy and reasoning consistency rate. We further manually analyzed the translation errors of Transformer on the entire test suite. We roughly grouped translation errors into three categories: common sense errors (translations that are not consistent with common sense), ordinary meaning errors (wrong translations of sources words that are not commonsense ambiguity points) and other errors (e.g., missing words). These errors were manually detected and labeled by two annotators. They checked all examples in the test suite. The inter-annotator agreement, measured as the rate of the number of labels that the two annotators annotate consistently against the total number of labels from the two annotators, is 92\%.

% \begin{table}
% \setlength{\belowcaptionskip}{-0.5cm}
% \centering
% \small
% \begin{tabular}{lcc}
% \toprule
% & RNNSearch & Transformer \\
% \midrule
% LA&0.24&0.26\\
% CL-SA &0.31&0.39 \\
% CT-SA &0.27&0.27\\
% Total&0.27&0.31\\
% \bottomrule
% \end{tabular}
% \caption
% {Rates of Reasoning consistency on the three test sets. }
% \label{table-5.6}
% \end{table}

Results are reported in Table \ref{table-5.9}. The majority of translation errors are indeed related to common sense (71.6\%). 
This suggests that our test suite is a suitable and challenging testbed for evaluating commonsense reasoning in NMT. 

\section{Conclusion}

In this paper, we have presented a test suite, including a lexical ambiguity test set and two syntactic ambiguity test sets, to evaluate the commonsense reasoning capability of state-of-the-art neural machine translation models.
We elaborate the rules of building this test suite and conduct statistical analyses on it.
Our evaluation experiments and analyses on this test suite suggest that commonsense reasoning in modern machine translation models is still in its infant stage and that more efforts are to be expected to advance NMT in this direction \cite{brown2020language,bras2020adversarial,ji2020language, long-etal-2020-ted}.

\section*{Acknowledgments}

The present research was supported by the National Natural Science Foundation of China (Grant No. 61861130364), Natural Science Foundation of Tianjin (Grant No. 19JCZDJC31400) and the Royal Society (London) (NAF$\backslash$R1$\backslash$180122). We would like to thank the anonymous reviewers for their insightful comments. The corresponding author is Deyi Xiong (dyxiong@tju.edu.cn).

\bibliography{emnlp2020}
\bibliographystyle{acl_natbib}

\end{CJK}

\end{document}